\documentclass{article} % For LaTeX2e
\usepackage{iclr2026_conference,times}

\usepackage{amsmath,amsfonts,bm}

\def\eqref#1{equation~\ref{#1}}
\def\1{\bm{1}}

\DeclareMathAlphabet{\mathsfit}{\encodingdefault}{\sfdefault}{m}{sl}
\SetMathAlphabet{\mathsfit}{bold}{\encodingdefault}{\sfdefault}{bx}{n}

\usepackage{hyperref}
\usepackage{url}

\usepackage{graphicx}
\usepackage{amsmath} 
\usepackage{amsfonts} 
\usepackage{multirow} 
\usepackage{subcaption} 
\usepackage{booktabs}
\usepackage{tabularx}
\usepackage{wrapfig}

\title{Reward-Forcing: Autoregressive Video Generation with Reward Feedback}

\author{
Jingran Zhang$^{1}$, Ning Li$^{1}$, Yuanhao Ban$^{2}$, Andrew Bai$^{2}$, Justin Cui$^{2}$ \\
$^{1}$University of California, San Diego 
$^{2}$University of California, Los Angeles \\
\texttt{\{jiz330, nil024\}@ucsd.edu, banyh2000@gmail.com} \\
\texttt{andrewbai@cs.ucla.edu, justincui@ucla.edu}
}

\iclrfinalcopy % Uncomment for camera-ready version, but NOT for submission.

\begin{document}

\maketitle

\begin{abstract}
While most prior work in video generation relies on bidirectional architectures, recent efforts have sought to adapt these models into autoregressive variants to support near real-time generation. However, such adaptations often depend heavily on teacher models, which can limit performance, particularly in the absence of a strong autoregressive teacher, resulting in output quality that typically lags behind their bidirectional counterparts. In this paper, we explore an alternative approach that uses reward signals to guide the generation process,  enabling more efficient and scalable autoregressive generation. By using reward signals to guide the model, our method simplifies training while preserving high visual fidelity and temporal consistency. Through extensive experiments on standard benchmarks, we find that our approach performs comparably to existing autoregressive models and, in some cases, surpasses similarly sized bidirectional models by avoiding constraints imposed by teacher architectures. For example, on VBench, our method achieves a total score of 84.92, closely matching state-of-the-art autoregressive methods that score 84.31 but require significant heterogeneous distillation.
\end{abstract}

\section{Introduction}
\label{sec:intro}

Diffusion models~\citep{ho2020denoising, liu2022flow} have emerged as a powerful class of generative models, achieving state-of-the-art results in a wide range of domains, including image synthesis, audio generation, and molecular modeling. By iteratively denoising data from a simple noise distribution, these models are capable of producing high-fidelity samples that capture complex data distributions. Their theoretical foundation rooted in stochastic differential equations, combined with their empirical robustness, has made diffusion models a dominant approach in the generative modeling landscape.

Building on this success, video diffusion models~\citep{ho2022video, videoworldsimulators2024} extend the capabilities of diffusion-based generation to the temporal domain. Unlike images, videos require the model to generate spatially coherent and temporally consistent sequences of frames, which significantly increases the modeling complexity. Recent advancements have adapted diffusion models to handle temporal correlations by incorporating spatiotemporal attention, motion priors, and multi-frame conditioning mechanisms. These models have demonstrated promising results in generating short to medium-length video clips with high perceptual quality.

However, most existing video diffusion models are built upon bidirectional architectures~\citep{wang2025wan, kong2024hunyuanvideo}, where the generation of each frame depends on information from both past and future timesteps. While effective for high quality video generation, bidirectional models are inherently unsuitable for streaming or real-time video generation, taking minutes to generate a single video with 5 seconds. Recent efforts have explored transforming bidirectional video diffusion models into autoregressive formulations~\citep{yin2025causvid, kim2025fifo} to enable low-latency and scalable generation. These methods typically rely on distillation from powerful bidirectional teachers, but their performance is often limited by the quality of the teacher model and the challenges of maintaining temporal coherence in the absence of future context.

In parallel, reinforcement learning (RL) has gained traction as a post-training strategy for enhancing the quality and alignment of diffusion-based image generation. By optimizing generation policies directly with respect to reward signals—such as human preferences, aesthetic scores, or perceptual quality—RL enables fine-grained control over generative behavior beyond supervised objectives. Techniques such as Reinforcement Learning with Human Feedback (RLHF)~\citep{ouyang2022training, liu2025improving} have demonstrated success in aligning language and vision-language models, and recent works have begun applying RL to tune diffusion models for improved fidelity, diversity, or alignment with desired styles.

In this work, we build upon these directions and present a novel approach that leverages autoregressive video diffusion modeling combined with reinforcement learning to achieve efficient, scalable, and high-quality streaming video generation. In summary, our contributions are as follows:
\begin{itemize}
    \item We show that the performance of existing methods for converting bidirectional video diffusion models into autoregressive models are bounded by the teacher's performance.
    \item We observe that consistent motions are learned before texture which aligns with previous studies on image generation and propose a framework that uses pure reward signals to guide  autoregressive generation of high quality videos.
    \item Extensive experiments on VBench show that our proposed method is able to generate high quality videos, comparable to baseline autoregressive video diffusion models on quality and total score.
\end{itemize}

\section{Related work}
\label{sec:related_work}

\paragraph{Diffusion Models for Image Generation} Denoising Diffusion Probabilistic Models (DDPM) pioneered the modern image generation by casting generative modeling as a Markovian noising–denoising process whose reverse dynamics are learned with a simple, noise‑conditional score network~\citep{ho2020denoising}. While DDPMs achieve impressive sample quality, they rely on hundreds or thousands of sequential denoising steps. Denoising Diffusion Implicit Models (DDIM) alleviate this inefficiency by interpreting the learned stochastic process as a non‑Markovian deterministic ordinary differential equation, enabling much faster inference without retraining and preserving visual fidelity~\citep{song2020denoising}. Subsequent flow‑matching methods such as the Flow Matching framework further unify diffusion and continuous‑normalizing‑flow viewpoints by directly training vector fields that map noise to data in a single pass, often reducing both training variance and sampling cost~\citep{lipman2022flow,liu2022flow}. The most widely used architecture for such tasks was UNet~\citep{ronneberger2015u}, which was later revolutionized by the introduction of DiT~\citep{peebles2023scalable}. Due to the high computational cost of operating directly in pixel space, modern architectures typically first project the input into a latent space~\citep{rombach2022high,blattmann2023stable}, where subsequent computations are performed. The final results are then decoded back into pixel space using a decoder~\citep{kingma2013auto}.

\begin{figure*}[t]
    \centering
    \includegraphics[width=\textwidth]{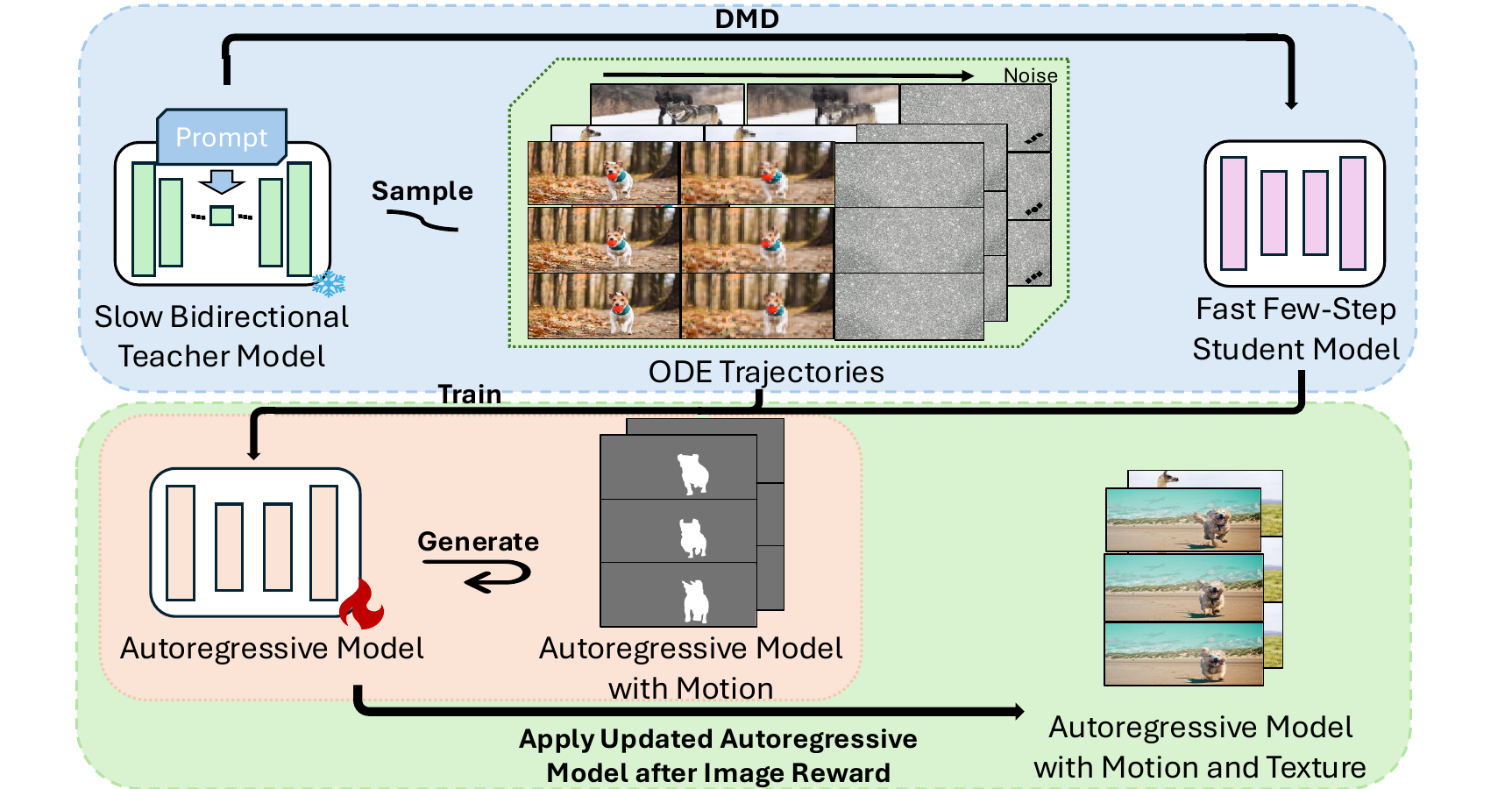}
    \caption{Overview of our proposed pipeline. The method first leverages a small set of ODE-based trajectories generated by a teacher model to guide the learning of motion dynamics. Subsequently, a reward model is employed to enhance the generation with fine-grained texture details.}
    \label{fig:overall-pipeline}
\end{figure*}
\paragraph{Video Diffusion Models}
Video diffusion models represent a significant advancement in generative AI, extending the principles of image diffusion models to handle temporal dynamics for tasks such as text-to-video generation, image-to-video synthesis, and video editing. Early extensions to video, such as Video Diffusion Models~\citep{ho2022video}, introduced 3D U-Net architectures with factorized spatio-temporal attention to ensure temporal coherence in video output with relative position embeddings~\citep{shaw2018self}. Subsequent models like Imagen Video~\citep{ho2022imagen} employed cascaded diffusion pipelines for high-definition videos, while Make-A-Video~\citep{singer2022make} leveraged pretrained text-to-image models and unsupervised video data to learn visual appearance and motion without requiring paired text-video data. By introducing novel spatial-temporal modules and a multi-stage video generation pipeline, it achieves state-of-the-art results in video quality, temporal consistency, and alignment with textual input. Latent space approaches, exemplified by models such as Stable Video Diffusion~\citep{blattmann2023stable,blattmann2023align}, enhanced efficiency by operating on compressed representations and incorporating temporal attention mechanisms. More recent innovations include zero-shot methods like Text2Video-Zero~\citep{khachatryan2023text2video}, which enable video generation without video-specific training, and Sora~\citep{openai2024chatgpt}, utilizing Diffusion Transformers on spacetime patches for scalable, high-fidelity outputs. These developments address challenges in temporal consistency and computational demands, paving the way for applications in entertainment, simulation, and content creation. The quality of generation is further enhanced by following models such as Wan~\citep{wan2025wan}, HunyuanVideo~\citep{kong2024hunyuanvideo} and Veo~\citep{veo}.

\paragraph{Diffusion Acceleration}
Although diffusion models are able to perform high quality image and video synthesis, they suffer from extremely high denoising steps with high cost. In order to solve the problem, one direction is to design better ODE solvers which enable few-step sampling such as DPM-Solver~\citep{lu2022dpm}, GENIE~\citep{dockhorn2022genie} and S4S~\citep{frankel2025s4s}. An alternative approach involves distillation techniques to compress multi-step diffusion processes into fewer or single-step generations, including Score Distillation Sampling (SDS) for leveraging pre-trained diffusion priors in optimization tasks~\citep{luo2024one}, Distribution Matching Distillation~\citep{yin2024one} which transforms diffusion models into efficient one-step generators with minimal quality loss, its enhanced successor DMD2~\citep{yin2024improved} that further improves training efficiency and performance, and Consistency Models that enable high-quality one- or few-step sampling by learning consistent mappings from noise to data~\citep{song2023consistency}.

\paragraph{Autoregressive and Streaming Generation}
While most diffusion models are trained with full-sequence denoising, they usually require substantially long time to generate videos up to several seconds. Thus autoregressive and streaming variants have been proposed to enable fast frame-by-frame or chunk by chunk generation ~\citep{yin2025causvid, huang2025self}. Among the work, FIFO-Diffusion~\citep{kim2025fifo} proposes a training-free method that uses a queue to hold frames of different noise levels with the frames closer to the beginning of the queue having lower noise level which requires less denoising and frames closer to the end of the queue have high noise levels which requires more denoising steps. The method is able to generate substantially longer videos than the original model. However, due to the heterogeneous denoising steps in the input, the generated videos often show degraded quality and inconsistency. Ouroboros-Diffusion~\citep{chen2025ouroboros} tries to solve the inconsistency problem by introducing a novel latent sampling technique at the end of the queue with subject-aware cross-frame attention mechanism. The heterogeneous noise level used by~\citet{kim2025fifo,chen2025ouroboros} shares the same idea as Diffusion-Forcing~\citep{chen2024diffusion,song2025history} where the models are trained to perform denoising on noisy frames with different noise levels. However, due to the large combination of noise levels, recent models~\citep{chen2025skyreels} usually first train the model with same noise level, then finetune it to handle different noise levels. Later work \citep{kodaira2025streamdit} tries to distill the model into few-step generators which alleviates the problem. Another line of works such as CausVid~\citep{yin2025causvid} and Self Forcing~\citep{huang2025self} adopts the same autoregressive manner but with same noise levels which show improved video quality.

\paragraph{Reinforcement Learning for Generative Models}
Reinforcement learning has been widely applied in Large Language Models~\cite{openai2024chatgpt,dubey2024llama,comanici2025gemini} to finetune  models to align better with user preferences~\cite{christiano2017deep,rafailov2023direct,liu2024deepseek}. Such techniques have also been generalized to diffusion models. ImageReward~\cite{xu2023imagereward} builds the first general-purpose text-to-image human preference reward model which can be used in the Reward Feedback Learning (ReFL) framework to effectively align image generation with human preferences. VisionReward~\cite{xu2024visionreward} extends it to both images and videos by designing a fine-grained, multi-dimensional reward model, achieving new state-of-the-art performances on various benchmark datasets. VideoAlign~\cite{liu2025improving} introduces a VLM-based reward model to address three critical dimensions including Visual Quality, Motion Quality and Text Alignment. ROCM~\cite{shekhar2025rocm} proposes a direct reward optimization framework for applying reinforcement learning from human feedback (RLHF) to consistency models, enabling efficient training without the need for policy gradients. Reward-Instruct~\cite{luo2025reward} achieves fast image synthesis through reward-centric approaches.

\section{Methodology}
\label{sec:methodology}
In this section, we will first introduce the background of diffusion models that distillation technique used for converting the model into a few step generator. Then we introduce how our method works. The overall pipeline is shown in Fig.~\ref{fig:overall-pipeline}.

\label{sec:background}
\subsection{Preliminaries}
\paragraph{Diffusion Model}
Diffusion‐based generative models treat data generation as running time backwards from a simple prior toward the data distribution. In the widely-used denoising diffusion probabilistic model (DDPM), one defines a forward Gaussian noise process

$$
\begin{aligned}
q\!\left(x_t \mid x_0 \right) &= \mathcal N\!\bigl(\alpha_t x_0,\;\sigma_t^{2}\mathbf I\bigr),\\
\alpha_t &= \prod_{s=1}^{t}(1-\beta_s)^{\tfrac12}, \quad \sigma_t^{2}=1-\alpha_t^{2}.
\end{aligned}
$$

where $x_0\sim p_{\text{data}}$ and $\{\beta_t\}_{t=1}^T$ is a small variance schedule. Generating a sample amounts to integrating the reverse-time score SDE
$$
\mathrm d x_t=\bigl[f(x_t,t)-g(t)^{2}\,\nabla_{x_t}\log q_t(x_t)\bigr]\mathrm d t+g(t)\,\mathrm d \bar w_t.
$$

or, in practice, its deterministic probability-flow ODE, using a neural score estimator $s_\theta(x,t)\approx\nabla_{x}\log q_t(x)$.

While DDPMs rely on stochastic trajectories, flow matching~\citep{lipman2022flow} learns a deterministic velocity field that continuously transports a tractable prior $p_0$ to the data distribution $p_1$.  Given a coupling $(x_0,x_1)\sim p_0\times p_1$, define linear interpolates $x_t=(1-t)\,x_0+t\,x_1$.  The ground-truth velocity $v^{\star}(x_t,t)$ is \[v^{\star}(x_t,t)=\frac{dx_t}{dt}=x_1-x_0.\]

and one trains a neural field $v_\theta(x,t)$ by the flow-matching loss

$$
\mathcal L(\theta)=\mathbb E_{t\sim\mathcal U(0,1)}\,
\mathbb E_{(x_0,x_1)}\Bigl[\,\bigl\|v_\theta(x_t,t)-v^{\star}(x_t,t)\bigr\|^{2}\Bigr].
$$

At inference, samples evolve deterministically via the ODE $\dot x_t=v_\theta(x_t,t)$, often requiring far fewer steps than stochastic diffusion samplers while maintaining high generative fidelity. In this work, we use Wan2.1~\citep{wang2025wan} which is trained using the flow matching objective.

\paragraph{Video Diffusion Distillation}
Same as CausVid~\citep{yin2025causvid} and SelfForcing~\citep{huang2025self}, we also utilize the distilled few-step model as the starting model for faster generation. It is achieved by minimizing the reverse KL divergence between data distribution and stduent generator's output distribution which can be formulated as\[
\begin{aligned}
\nabla_{\theta}\mathcal{L}_{\mathrm{DMD}}
&= \mathbb{E}_{t}\!\bigl[
      \nabla_{\theta}\operatorname{KL}\bigl(
         p_{\mathrm{fake},t}\Vert p_{\mathrm{real},t}
      \bigr)
   \bigr] \\[2pt]
&= -\,\mathbb{E}_{t}\!\Bigl[
      \int\!\bigl(
         s_{\mathrm{real}}\bigl(\Phi(G_{\theta}(z),t),t\bigr)
   \\[-2pt]
&\qquad\;\;
         -\,s_{\mathrm{fake}}\bigl(\Phi(G_{\theta}(z),t),t\bigr)
      \bigr)\,
      \frac{dG_{\theta}(z)}{d\theta}\,dz
   \Bigr].
\end{aligned}
\]

where $\Phi$ is the forward diffusion process and $z \sim \mathcal{N}(0,\mathbf{I})$ is a random Gaussian noise input.

\subsection{Training with ODE and Reward Guidance}
Recent studies have shown that diffusion models typically generate a global coarse structure before progressively refining texture details~\citep{chen2023motion,jiang2024poetry2image,sun2025ds,liu2025detailflow}. We empirically observe a similar phenomenon when converting bidirectional video diffusion teachers into autoregressive models which can be seen in Fig.~\ref{fig:two-rows-eight-images} here. After distilling the model into few-step models, similar to CausVid~\citep{yin2025causvid}, we use sample ODE trajectories for early training. The process is done by first sample noise inputs $\{x_T^i\}_{i=1}^L$ from $\mathcal{N}(0, I)$ and then use an ODE solver with the pretrained teacher to generate reverse trajectories $\{x_t^i\}_{i=1}^L$ across all timesteps. The model is then trained with the following objective:
$$
\mathcal{L}_{\text{ode}} = \mathbb{E}_{x, t^i} \left\| G_\phi(\{x_{t^i}^i\}, \{t^i\}) - \{x_0^i\} \right\|^2.
$$
where $G_\phi$ is the student trained from the teacher. After this step, we empirically observe that the model has learned to generate consistent motions without much texture info.

\paragraph{Optimization with Reward Feedback}
After initializing the model with an ODE-based process that teaches motion synthesis, we introduce a reward-guided optimization stage to enhance video quality. Specifically, we adopt ImageReward~\citep{xu2023imagereward} as our reward model and incorporate it into training in a differentiable manner to directly guide the video diffusion model.

Let the generated video be denoted as 
$$\hat{x}_{1:T} = G_\theta(z)$$, 
where $G_\theta$ is the autoregressive video generator parameterized by $\theta$, and $z$ is the latent input. The reward model $\mathcal{R}(\cdot)$ assigns a scalar reward indicating perceptual quality. We define the reward-guided objective as:
$$
\mathcal{L}_{\text{reward}}(\theta) = -\mathbb{E}_{z \sim \mathcal{Z}} \left[ \mathcal{R}(\hat{x}_T) \right].
\label{eq:reward_loss}
$$

where $\hat{x}_T$ is the last frame of the generated video. We choose to apply supervision to the last frame because we find that supervising more frames encourages static content with reduced motion, while supervising the last frame preserves motion better likely due to its proximity to the end of the generation trajectory in the autoregressive process. Results of supervising random frames are shown in ablation study.

\subsection{Baseline Methods}
We benchmark three classes of approaches: (i) standard bidirectional diffusion models, (ii) autoregressive methods that render videos one (latent) frame at a time, and (iii) autoregressive methods that synthesize videos in temporal chunks. For standard diffusion models, we include LTX-Video~\citep{hacohen2024ltx} and Wan2.1~\citep{wan2025wan} which is the base model used by~\citet{yin2025causvid, huang2025self} and our method. For frame-wise autoregressive models, we include NOVA~\citep{deng2024nova} which formulates the video generation problem as non-quantized autoregressive modeling of temporal frame-by-frame prediction and spatial set-by-set prediction, Pyramid Flow~\citep{jin2024pyramidal} that interprets the original denoising trajectory as a series of pyramid stages and Self Forcing~\citep{huang2025self} that trains the autoregressive models by relying on self-generated frames instead of ground-truth ones. For Chunk-wise models, we include SkyReels-V2~\citep{chen2025skyreels}, MAGI-1\citep{magi1} that utilizes transformer based VAE and increased noise levels among chunks that are being generated, CausVid~\citep{yin2025causvid} that uses causal attention at training time and self-rollout at inference time and Self Forcing~\citep{huang2025self} which utilizes similar architectures as its frame-wise counterpart.

\begin{table*}[t]
  \small
  \centering
  \resizebox{\textwidth}{!}{
      \begin{tabular}{lccccccc}
          \toprule
          \multirow{2}{*}{Model} & \multirow{2}{*}{\#Params} & \multirow{2}{*}{Resolution} & \multirow{2}{*}{Throughput} & \multirow{2}{*}{Latency} &
          \multicolumn{3}{c}{Evaluation scores $\uparrow$}\\
        \cmidrule(lr){6-8}
         & & & (FPS) $\uparrow$ & (s) $\downarrow$ & Total & Quality & Semantic \\
         & & &                  &                  & Score & Score & Score \\
        \midrule
        % \midrule
        % \rowcolor{catgray}
        \multicolumn{8}{l}{\textit{Diffusion models}}\\
        LTX-Video      & 1.9B & $768{\times}512$ & 8.98          & 13.5          & 80.00 & 82.30 & 70.79 \\
        Wan2.1           & 1.3B & $832{\times}480$ & 0.78          & 103           & 84.26 & 85.30 & 80.09 \\
        \midrule
        % \rowcolor{catgray}
        \multicolumn{8}{l}{\textit{Frame-wise Autoregressive models}}\\
        NOVA             & 0.6B & $768{\times}480$ & 0.88          & 4.1  & 80.12 & 80.39 & 79.05 \\
        Pyramid Flow & 2B   & $640{\times}384$ & 6.7           & 2.5           & 81.72 & 84.74 & 69.62 \\
        Self Forcing (frame-wise)      & 1.3B & $832{\times}480$ & 8.9           & 0.45 & 84.26 & 85.25 & 80.30 \\ % 8.9, 0.45
        \midrule
        % \rowcolor{catgray}
        \multicolumn{8}{l}{\textit{Chunk-wise autoregressive models}}\\
        SkyReels-V2  & 1.3B & $960{\times}540$ & 0.49          & 112           & 82.67 & 84.70 & 74.53 \\
        MAGI-1                 & 4.5B & $832{\times}480$ & 0.19          & 282           & 79.18 & 82.04 & 67.74 \\
        CausVid  & 1.3B & $832{\times}480$ & 17.0 & 0.69          & 81.20 & 84.05 & 69.80 \\
        Self Forcing      & 1.3B & $832{\times}480$ & 17.0 & 0.69          & 84.31 & 85.07 & \textbf{81.28}    \\  % 17.0, 0.69
        \midrule
        ODE Only (chunk-wise)      & 1.3B & $832{\times}480$ & 17.0           & 0.69 & 68.77 & 73.27 & 50.81 \\ % 8.9, 0.45
        Ours (chunk-wise)      & 1.3B & $832{\times}480$ & 17.0           & 0.69 & \textbf{84.92} & \textbf{85.91} & 80.97 \\ % 8.9, 0.45
        % CausForce (Ours, Reward+Distillation)      & 1.3B & $832{\times}480$ & 17.0           & 0.69 & 84.79 & 85.73 & 81.03 \\ % 8.9, 0.45
        \bottomrule
      \end{tabular}}\\
  \caption{
    Overall performance comparison with previous baseline methods using VBench. 
  }
    \label{tab:main}

  %\vspace{2pt}
  %\vspace{-1em}
\end{table*}

\subsection{Implementation Details}
Similar to~\citet{yin2025causvid, huang2025self}, our model is based on Wan2.1-T2V-1.3B\citep{wan2025wan}. The model is first distilled into a 4-step model using DMD. For ODE initialization, we also directly sample 1.4K trajectories from the bidirectional model. Our work uses Self-Rollout during training which is the same as SelfForcing~\citep{huang2025self}. For our main method, we only employ loss from the reward models. For our reward plus distillation setting, we train the model using DMD together with reward loss as shown in the previous section.  Our method also works for both chunk-wise and frame-wise autoregressive generation. In our implementation, since we only need to sample the ODE trajectory from the teacher model, we don't need any real training dataset. Thus our overall approach remains data free which is similar to that of Self Forcing~\citep{huang2025self}. Similar to previous works~\citep{huang2025self}, we enable EMA~\citep{hunter1986exponentially} during the training process.

Note that our training process eliminates the second distillation stage used in CausVid~\citep{yin2025causvid} and Self Forcing~\citep{huang2025self}. As a result, there is no need to load the teacher and critic models or train the critic model to approximate the generator distribution, making the training process after ODE initialization significantly more lightweight and efficient.

\section{Experimental Results}
\label{sec:experimental_results}

\subsection{Evaluation Metrics}
We evaluate our method on VBench~\citep{huang2024vbench}, following the protocol in~\cite{yin2025causvid, huang2025self}. VBench comprises 16 evaluation dimensions, which are aggregated into three scores: quality, semantic, and total. The quality score is a weighted average of metrics such as subject consistency, background consistency, temporal flickering, motion smoothness, aesthetic quality, and dynamic degree, reflecting the technical fidelity of generated videos. The semantic score averages dimensions like object class, multiple objects, human action, color, spatial relationships, and overall consistency, capturing content relevance and semantic coherence. Since we adopt a similar KV cache strategy, our inference runtime matches that of Self Forcing. In addition to the aggregated scores, we also report per-dimension results to provide a more fine-grained comparison of visual quality and semantic consistency across different methods.

\subsection{Main Results}

\setlength{\intextsep}{3pt}      % space above/below wrapfig
\setlength{\columnsep}{10pt}     % space between text and figure

\begin{wrapfigure}{r}{0.47\linewidth}
    \centering
    \includegraphics[width=\linewidth]{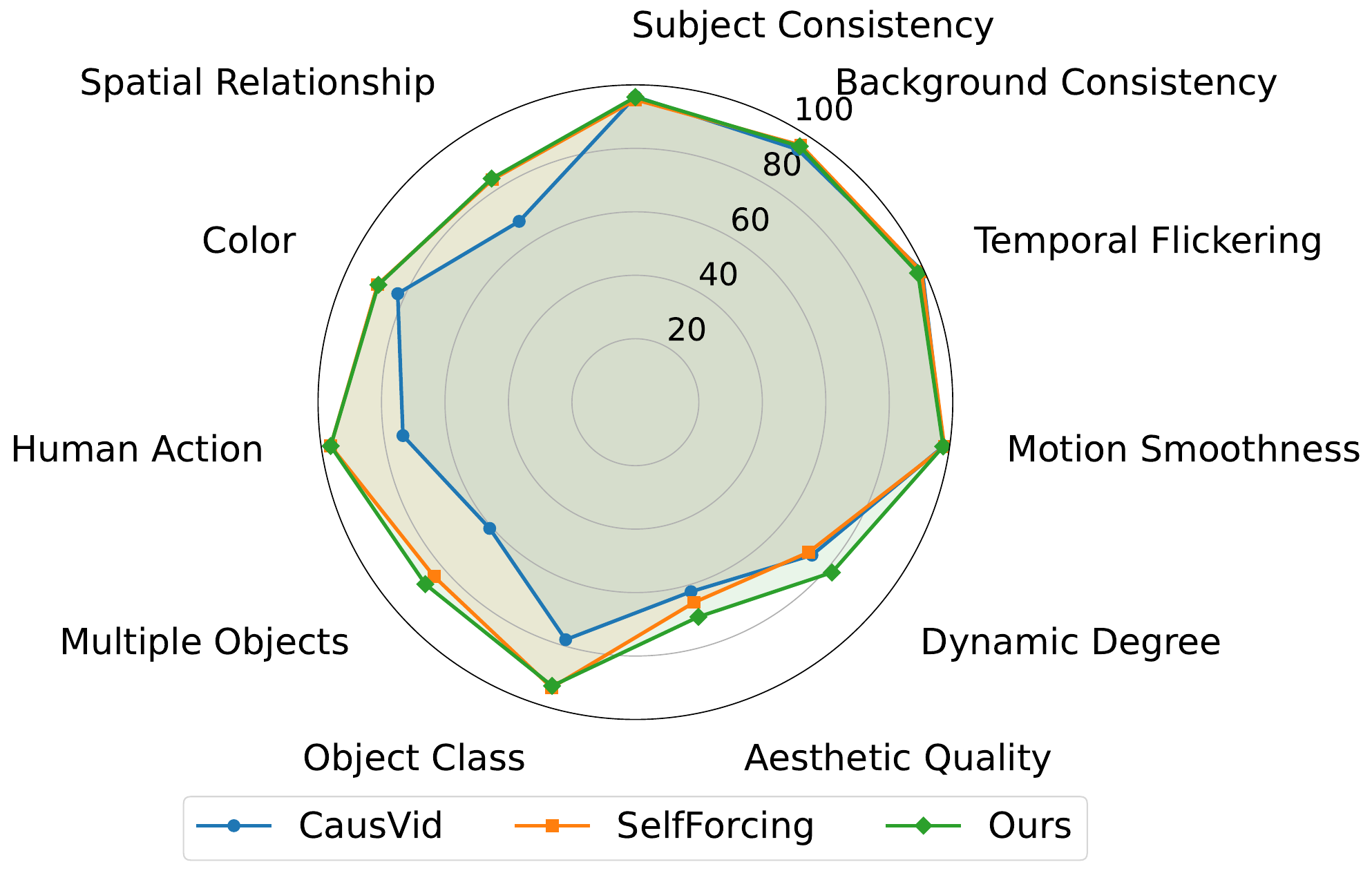}
    \caption{Comparison between our methods and baseline methods on selected VBench metrics. Our method shows competitive performances without extensive heterogeneous distillation.}
    \label{fig:vbench-plot}
\end{wrapfigure}
\par

Following prior works~\citep{yin2025causvid, huang2025self}, we report aggregated scores in Tab.~\ref{tab:main}, with selected metrics plot in Fig.~\ref{fig:vbench-plot}.

Overall, standard bidirectional teacher models tend to perform better than autoregressive models in many aspects. For instance, the bidirectional model Wan2.1~\citep{wan2025wan} achieves a total score of 84.26, whereas CausVid~\citep{yin2025causvid} attains a lower score of 81.20, likely due to architectural differences and a mismatch between training and inference procedures. Self Forcing~\citep{huang2025self} attempts to address this issue by incorporating self-rollout during training, which helps reduce the discrepancy and leads to improved results. However, since it is distilled from the teacher using DMD~\citep{yin2024improved}, its performance remains largely comparable to that of the bidirectional model with same size.

Our method, despite being relatively simple and not relying on extensive heterogeneous distillation between bidirectional teacher and autoregressive student model, achieves competitive results and shows encouraging signs in comparison to both the bidirectional model and recent autoregressive baselines. For example, it achieves a quality score of 85.81, slightly higher than the 85.25 obtained by state-of-the-art frame-wise Self Forcing. It also reaches a total score of 84.92, outperforming the second-best score achieved by the chunk-wise Self Forcing model. As shown in Fig.~\ref{fig:vbench-plot}, our method performs favorably on several individual dimensions, such as aesthetic quality and dynamic degree.

One interesting observation is that as the distillation process progresses, the dynamic degree of motion tends to diminish. While more analysis is needed, this may suggest a trade-off between distillation and motion richness, which we hope to explore further in future work. Additionally, although our model is only initialized with a small number of ODE trajectories, Tab.~\ref{tab:main} shows meaningful improvements over the initial model in both quality and semantic scores. Taken together, these findings suggest that with careful design, autoregressive models may be able to close the performance gap with teacher models, possibly even without relying heavily on distillation. We believe this points toward a promising direction worth further investigation.

\subsection{Qualitative Comparison}
\begin{figure*}[t]
    \centering
    \includegraphics[width=\textwidth]{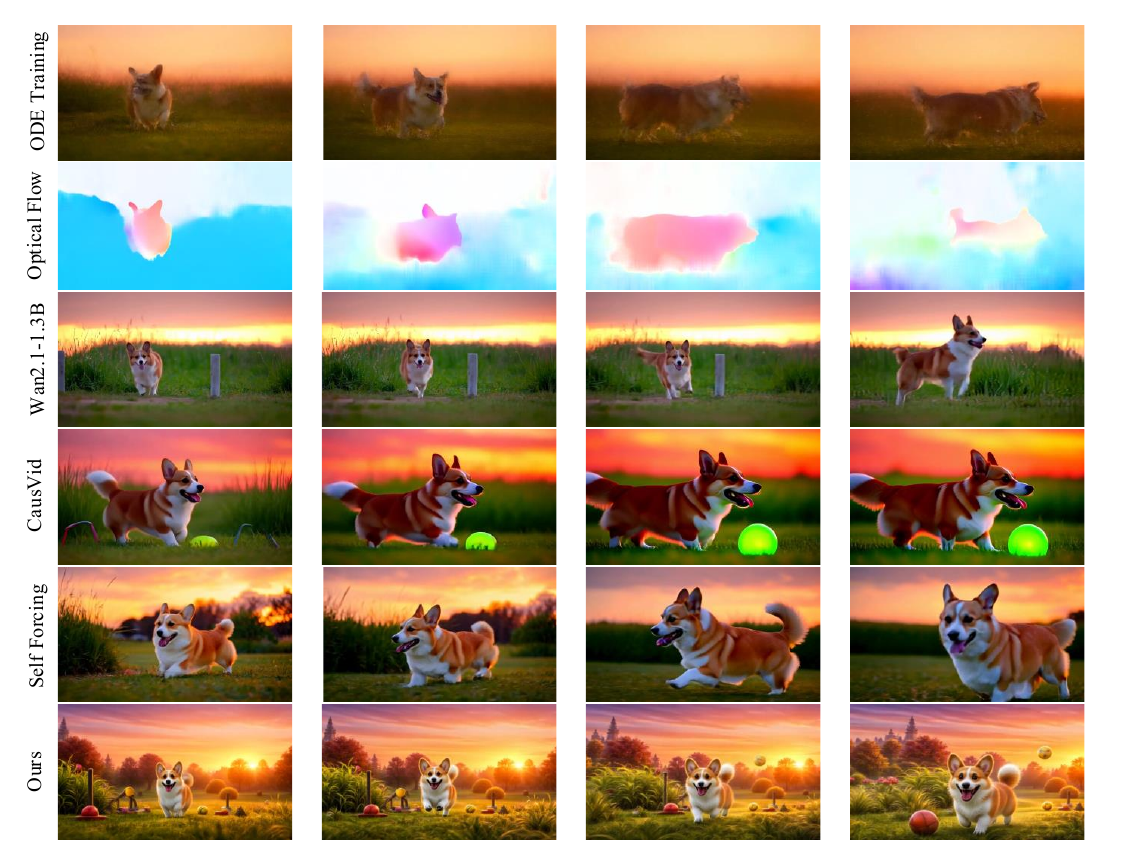}
    \caption{Comparison of videos generated by our method and other baseline methods, using the prompt: ``A joyful, playful Corgi running and frolicking in a vibrant park during sunset.'' The second row displays the optical flow of the video generated by the model trained solely with ODE trajectories, highlighting the motion patterns the model has learned.}
    \label{fig:two-rows-eight-images}
\end{figure*}

Here we show the generated videos in Fig.~\ref{fig:two-rows-eight-images} and compare with two other state-of-the-art models including CausVid~\citep{yin2025causvid} and Self Forcing~\citep{huang2025self} and the original Wan2.1 1.3B model. It can be seen that our model is able to generate high quality videos compared to the baseline model. Note that since the texture of our videos are generated with the guidance of an external reward model, the overall style will look different with that of the teacher and baseline models.

\vspace{-5px}

\section{Ablation Study}
\label{sec:ablation_study}

\subsection{Combining Reward with Distillation}
One alternative we explored is to directly combine the reward loss with the distribution matching loss and jointly optimize both objectives during training. While this approach seems intuitive, balancing the teacher's guidance with task-specific reward supervision, we observed that it actually results in degraded performance. Specifically, when training under a comparable computational budget, the final model only achieves a total score of 82.55, which is notably lower than both the original SelfForcing baseline and our proposed method. We believe this performance drop is rooted in a fundamental conflict between the objectives of distribution matching and reward-based optimization. The distillation framework is explicitly designed to align the output distribution of the student model with that of the teacher, ensuring that the student faithfully mimics the teacher’s behavior. However, by introducing a reward loss, especially one derived from models such as ImageReward, the student is incentivized to generate outputs that may diverge from those of the teacher in pursuit of higher reward scores. This divergence undermines the distribution alignment objective, weakening the distillation process and leading to overall lower performance. These results highlight the importance of carefully decoupling teacher supervision from reward optimization. The results can be found in Tab.~\ref{tab:scores_summary} 
\begin{table}[h]
\centering
\small
\resizebox{\textwidth}{!}{%
\setlength{\tabcolsep}{24pt}
\begin{tabular}{lccc}
\toprule
\textbf{Model} & \textbf{Total Score} & \textbf{Quality Score} & \textbf{Semantic Score} \\
\midrule
CausVid         & 81.20 & 84.05 & 69.80 \\
Self Forcing     & 84.31 & 85.07 & \textbf{81.28} \\
Reward + Distill    & 82.55 & 83.20 & 79.96 \\
Ours         & \textbf{84.92} & \textbf{85.91} & 80.97 \\
\bottomrule
\end{tabular}}
\caption{Performance comparison between our method and one combining both reward and distillation losses. Results are evaluated under similar compute. No further gains observed with more computation}
\label{tab:scores_summary}
\end{table}

\subsection{Random Steps vs Last Step}
Since a video is inherently a temporal sequence of image frames, there are multiple strategies for applying reward models during training. Common approaches include: (1) applying the reward only to the last frame of the video versus (2) applying the reward to randomly selected frames throughout the sequence. In our experiments, we empirically find that supervising randomly selected frames results in a degradation in motion quality, with the motion degree dropping by more than 10 percentage points. We believe this decline is primarily due to the nature of the ImageReward model, which is designed to evaluate individual frames based on visual-textual alignment and focuses primarily on texture-level details. Because it lacks any notion of temporal continuity or cross-frame motion dynamics, it fails to capture the quality of motion across frames. As a result, applying it to random frames can encourage the model to prioritize static visual features over coherent motion, ultimately harming the motion of the generated videos.

\subsection{Different Teacher Models}
Previously, we observe that incorporating both distillation and reward loss into the training objective does not necessarily improve performance. This raises the question of whether using a different teacher model could yield different results. We experiment with two teacher configurations: the larger Wan2.1-14B model and the smaller Wan2.1-1.3B model, both combined with the same image-based reward loss to guide the student. Across multiple runs and evaluation metrics, we do not observe substantial differences in performance. We hypothesize that this is due to the limited capacity of the 1.3B student model, which may reach its performance ceiling regardless of their size or expressiveness. Additionally, the fundamental difference between the bidirectional teacher and the autoregressive student model may also contribute to the performance gap.

\section{Conclusion \& Future Work}
\label{sec:conclusion}

In this work, we introduce an alternative approach for converting bidirectional video diffusion models into autoregressive counterparts by leveraging the guidance of reward models. Unlike prior methods that rely heavily on heterogeneous distillation from pre-trained bidirectional teachers, our framework eliminates the need for this additional training stage. We show that it is feasible to initialize the autoregressive model directly using the ODE-based training of the original diffusion model, and then progressively refine its generation capabilities through reward-guided optimization. By avoiding the constraints of strict teacher-student alignment, our method offers the potential to unlock more expressive and higher-quality video generation.

One promising direction for future research is to explore alternative reward models, including both image- and video-based rewards, to further improve the quality of generated videos. Our current approach builds on the findings ~\citep{jiang2024poetry2image,sun2025ds,liu2025detailflow,materzynska2024newmove}, which suggest that global structures in images and motions in videos tend to emerge early in the generative process. Directly modeling motion dynamics could potentially further enhance generation quality~\citep{shi2024motion,chefer2025videojam}. In addition, Reward-Instruct~\citep{luo2025reward} shows that high-quality image synthesis can be achieved solely through reward-guided training while bypassing the heavy distillation process. Inspired by this, future work could investigate the possibility of eliminating the initial ODE training phase entirely in favor of a reward-driven learning paradigm for video generation.

\section{Limitations}
\label{sec:limitations}
Although our method demonstrates competitive performance relative to baseline approaches, video generation inherently involves more complex and multi-dimensional judgments than image generation. Our framework relies entirely on open-source reward models, and users may select reward functions that best suit their downstream applications. We propose an alternative pathway for converting bidirectional teacher models into autoregressive student models; however, since bidirectional video diffusion models already provide strong guidance, we do not apply our method to standard non-autoregressive settings, which we leave for future work.  Due to limited teacher supervision, we observe occasional inconsistencies in generated videos, which could potentially be mitigated by stronger or more expressive reward models. From a practical standpoint, improving the stability and efficiency of reward-based training for large-scale video models—through better optimization strategies, variance reduction, and scalable reward modeling—remains an important direction for future research.

\bibliography{iclr2026_conference}
\bibliographystyle{iclr2026_conference}

\clearpage
\appendix
\section{Appendix}

\subsection{More related works}
\paragraph{Motion in Video Diffusion Models}
Motion modeling remains a central challenge in video diffusion. Many early video diffusion models, such as Video Diffusion Models \citep{ho2022video} primarily focus on spatiotemporal attention mechanisms but treat motion implicitly. More recent efforts like MCDiff \citep{chen2023motion} and VideoJAM \citep{chefer2025videojam} attempt to model motion more explicitly, using a flow completion model or joint appearance-motion representations. And Motion-I2V \citep{shi2024motion} further explores motion disentanglement in the image-to-video task, suggesting a broader trend toward treating motion and appearance as separable generative components.

\section{Implementation Details}

\subsection{Computing Infrastructure}
All experiments were conducted on 8 H100 GPUs with 80 GB of memory, running on a Linux operating system. The versions of all relevant libraries and frameworks will be detailed in the later released GitHub repository.

\subsection{Hyperparameters}
Here we list the hyperparameters we used for our experiment. We use a batch size of 8 after training with the ODE trajectory. For the optimizer, we use AdamW with $\beta_1=0,\beta_2=0.999, \epsilon=1e-8,weight\_decay=0.01$ with a learning rate of $2e-6$. For EMA, we use a decay of 0.99. When combing distillation loss with image reward loss, we normalize the losses so that they are on the same scale. When comparing with baseline methdos such as DMD or Causvid, we use their author-provided checkpoints.

\section{Quality of Generated Videos}
It is important to recognize that our autoregressive model has not been fully distilled from the bidirectional teacher model, which processes information from both directions for a holistic view. Consequently, the videos generated by our model may differ substantially from those of the bidirectional teacher. These trajectories guide the video's temporal dynamics, and deviations could lead to unique visual patterns or flows.

Additionally, our model's texture generation is directed by a reward model, which provides evaluative feedback to align outputs with desired criteria. This allows downstream users to select or customize their own reward models, enabling the creation of videos tailored to specific styles, such as artistic animations or realistic footage, enhancing personalization.

This design also means our final model's performance is not strictly limited by the teacher. Instead, the autoregressive model can potentially outperform the teacher, guided by reward models, in areas like motion quality (e.g., smoother transitions) or aesthetic quality (e.g., richer details and harmony) if strong reward models are used.

\section{Generalization to Long Videos}
Since our model is directly built upon CausVid~\citep{yin2025causvid} and Self Forcing~\citep{huang2025self}, thus the model can be used to generate long videos as well. E.g. Causvid generates long videos by conditioning on previous generated frames which could be utilized to generate long videos by our method too.

\section{Full Evaluation Metrics}
Tab. \ref{tab:full_vbench} presents the comprehensive evaluation across all 16 dimensions of VBench. The results show our method maintains competitive performance across most metrics while exhibiting particular strengths in certain areas. Our approach achieves a dynamic degree of 81.94, suggesting the ODE-based motion initialization contributes to motion representation. We also observe an aesthetic quality score of 70.51, aligning with the reward-guided refinement strategy. Additionally, the method yields scores of 87.58 in multiple objects and 83.78 in spatial relationship metrics, indicating its capability in preserving compositional relationships.

\begin{table}[t]
  \centering
  \resizebox{\textwidth}{!}{%
  \setlength{\tabcolsep}{24pt}
  \begin{tabular}{lccc}
    \toprule
    \multirow{2}{*}{Metric} & \multicolumn{3}{c}{Score $\uparrow$} \\
    \cmidrule(lr){2-4}
    & CausVid & Self Forcing & Ours \\
    \midrule
    Subject Consistency      & 96.32 & 95.32 & 96.15 \\
    Background Consistency   & 94.65 & 96.30 & 95.83 \\
    Temporal Flickering      & 99.38 & 99.05 & 97.88 \\
    Motion Smoothness        & 98.00 & 98.35 & 97.99 \\
    Dynamic Degree           & 73.61 & 72.22 & 81.94 \\
    Aesthetic Quality        & 62.24 & 65.75 & 70.51 \\
    Imaging Quality          & 67.82 & 69.02 & 68.84 \\
    Object Class             & 78.01 & 93.75 & 93.20 \\
    Multiple Objects         & 60.75 & 83.92 & 87.58 \\
    Human Action             & 74.00 & 97.00 & 97.00 \\
    Color                    & 82.34 & 89.28 & 89.04 \\
    Spatial Relationship     & 67.74 & 83.38 & 83.78 \\
    Scene                    & 14.03 & 57.34 & 54.36 \\
    Appearance Style         & 21.23 & 24.40 & 24.06 \\
    Temporal Style           & 18.85 & 20.48 & 21.19 \\
    Overall Consistency      & 22.50 & 26.72 & 26.14 \\
    \bottomrule
  \end{tabular}}
  \caption{Complete VBench evaluation metrics in all 16 dimensions. Our method shows improvements in motion-related metrics (Dynamic Degree) and perceptual quality (Aesthetic Quality), while maintaining competitive performance on other dimensions.}
  \label{tab:full_vbench}
\end{table}

\subsection{More Generated Samples}

\begin{figure*}[t]
    \centering
    \begin{subfigure}[b]{\textwidth}
        \centering
        \includegraphics[width=0.23\linewidth]{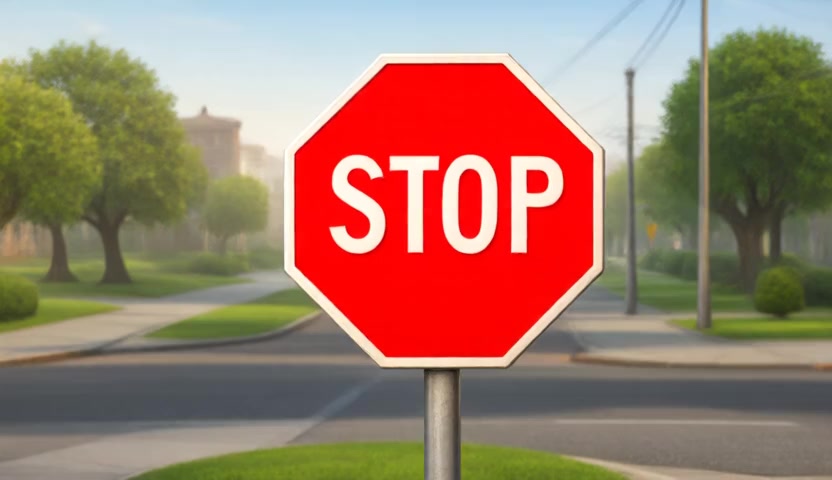}
    \hfill
        \includegraphics[width=0.23\linewidth]{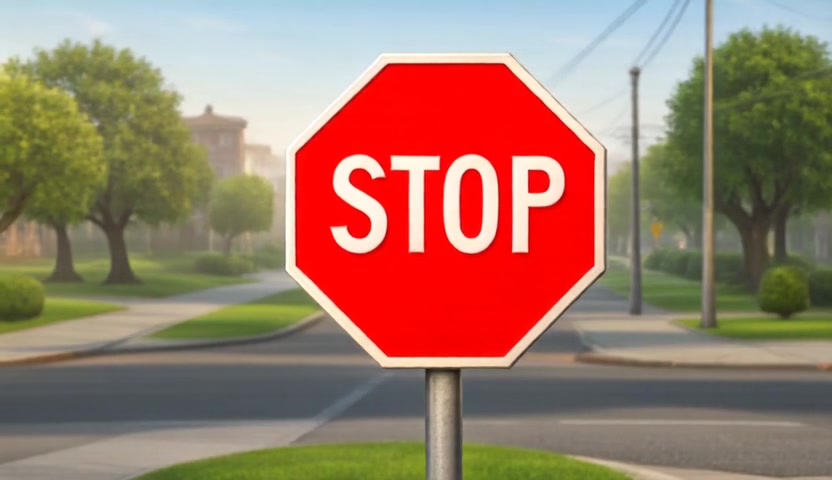}
    \hfill
        \includegraphics[width=0.23\linewidth]{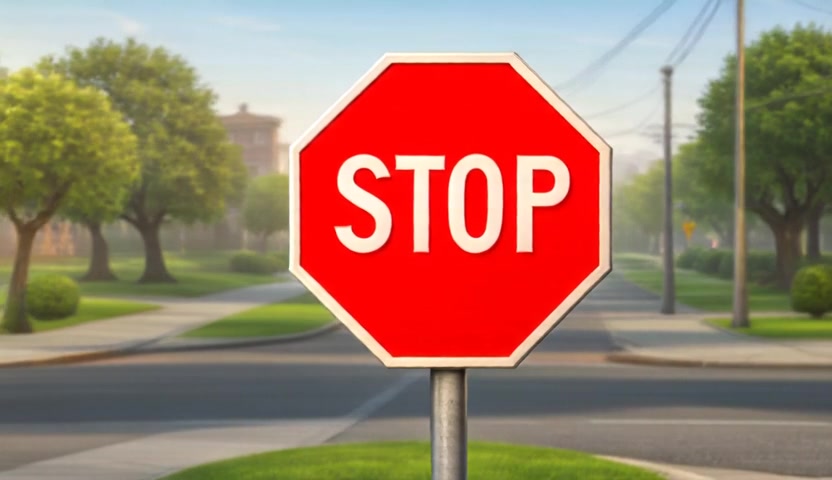}
    \hfill
    \includegraphics[width=0.23\linewidth]{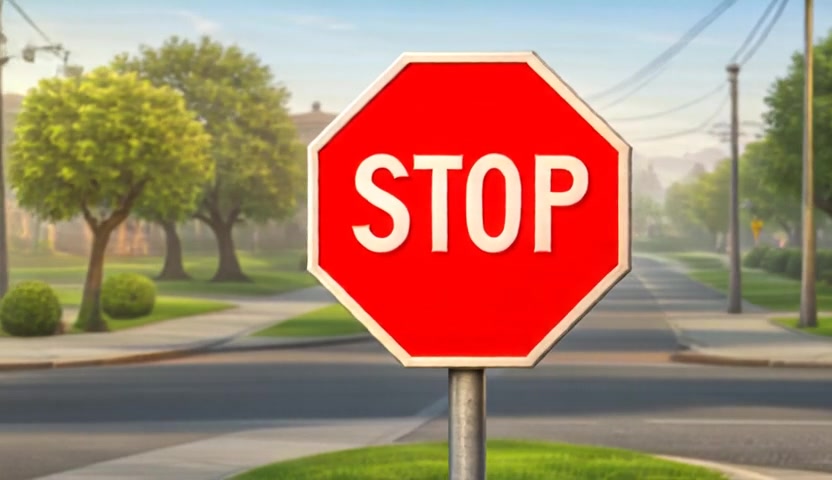}
    \subcaption{In a still frame, a stop sign}
    \end{subfigure}

        \begin{subfigure}[b]{\textwidth}
        \centering
        \includegraphics[width=0.23\linewidth]{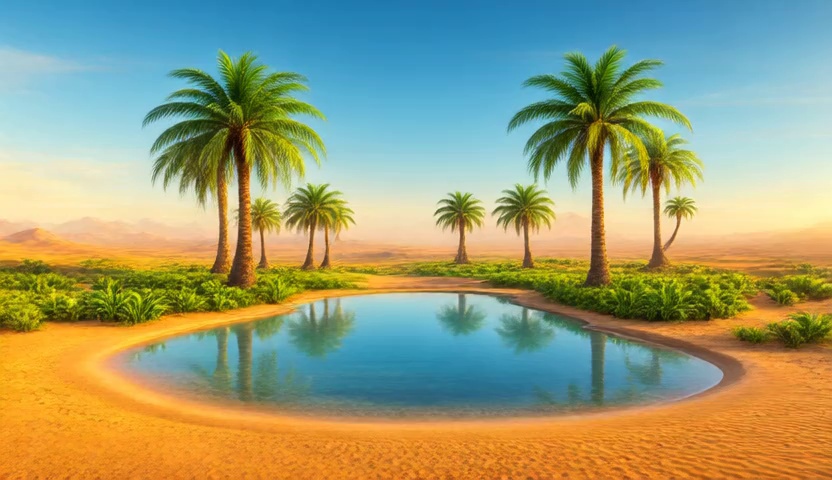}
    \hfill
        \includegraphics[width=0.23\linewidth]{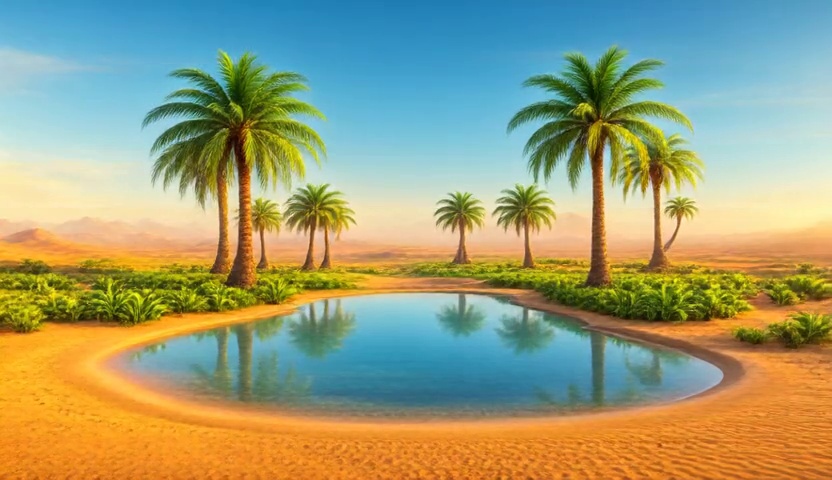}
    \hfill
        \includegraphics[width=0.23\linewidth]{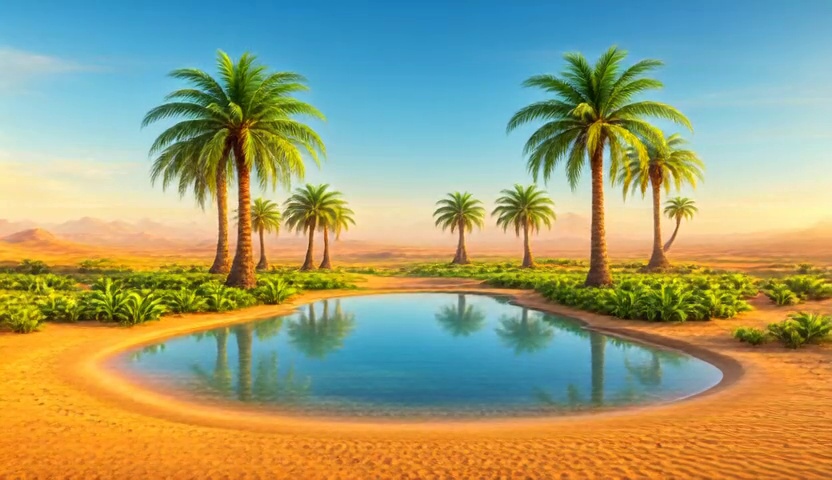}
    \hfill
    \includegraphics[width=0.23\linewidth]{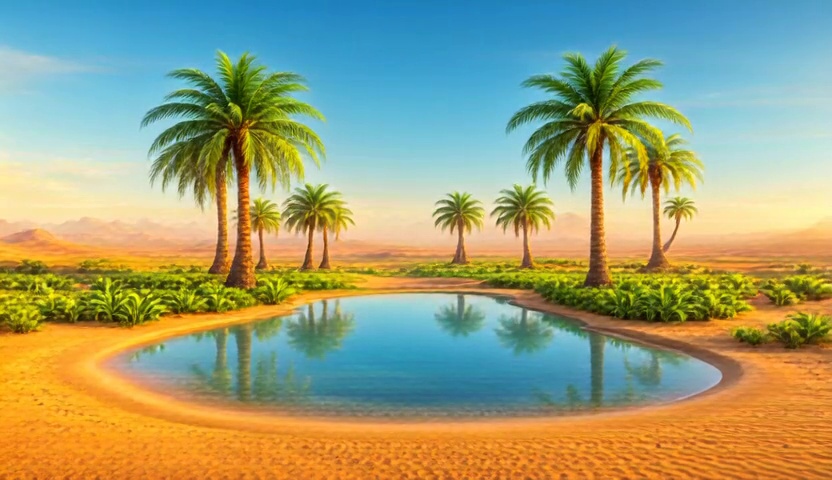}
    \subcaption{In a still frame, within the desolate desert, an oasis unfolded, characterized by the stoic presence of palm trees and a motionless, glassy pool of water}
    \end{subfigure}

\begin{subfigure}[b]{\textwidth}
        \centering
        \includegraphics[width=0.23\linewidth]{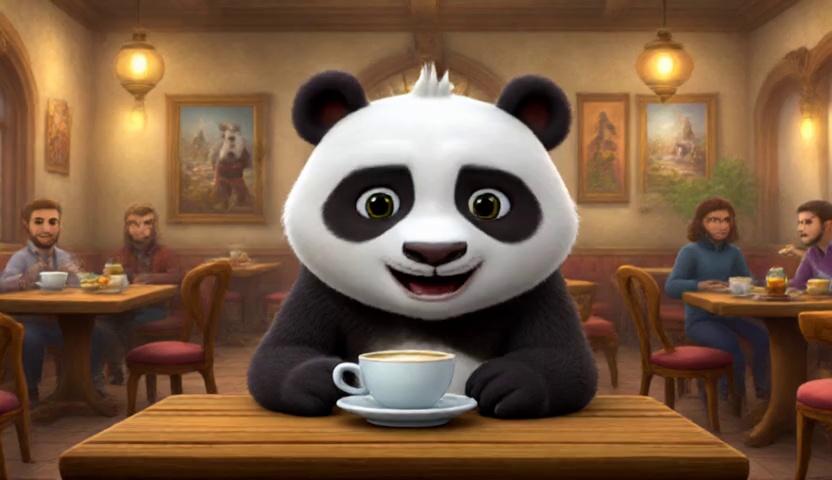}
    \hfill
        \includegraphics[width=0.23\linewidth]{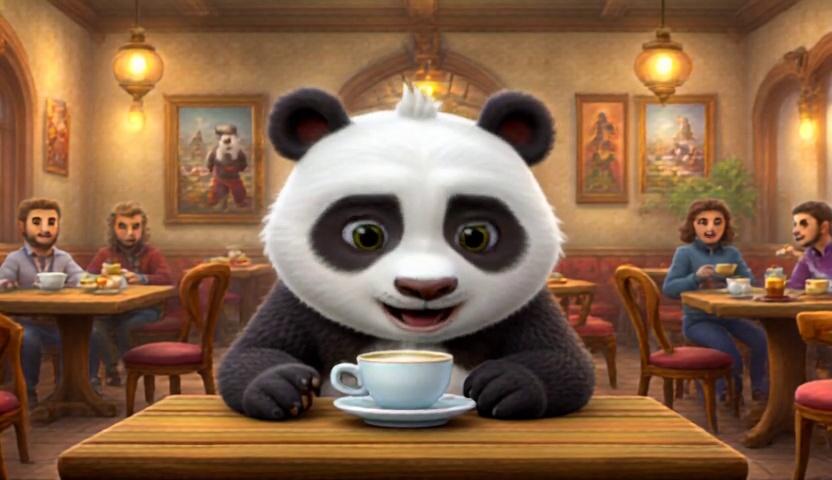}
    \hfill
        \includegraphics[width=0.23\linewidth]{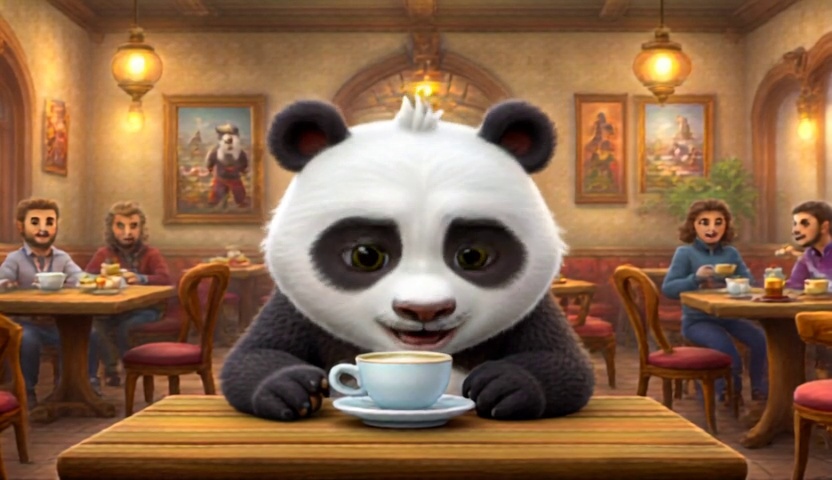}
    \hfill
    \includegraphics[width=0.23\linewidth]{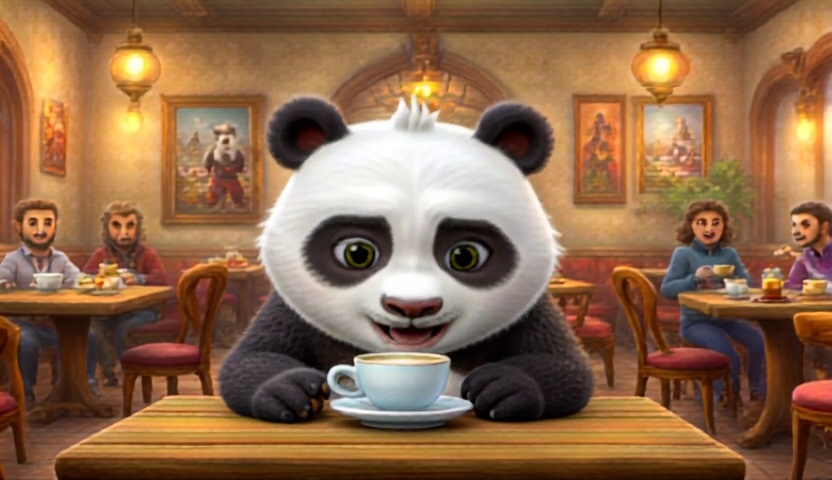}
    \subcaption{A panda drinking coffee in a cafe in Paris, featuring a steady and smooth perspective}
    \end{subfigure}

    \begin{subfigure}[b]{\textwidth}
        \centering
        \includegraphics[width=0.23\linewidth]{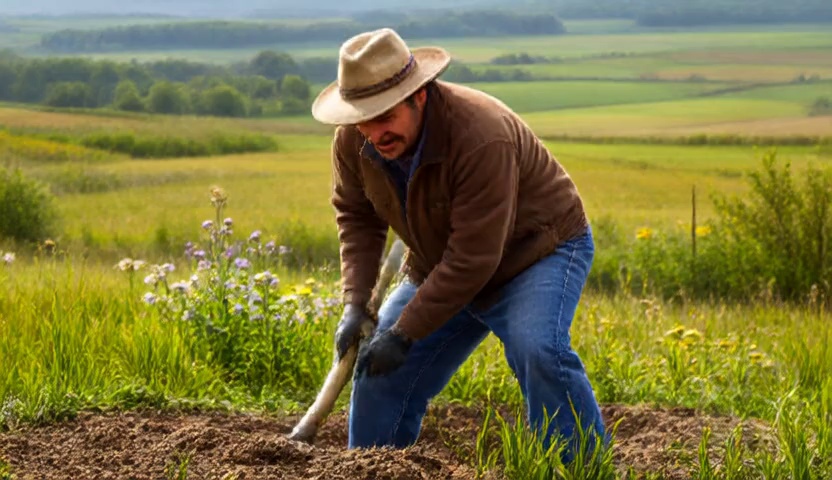}
    \hfill
        \includegraphics[width=0.23\linewidth]{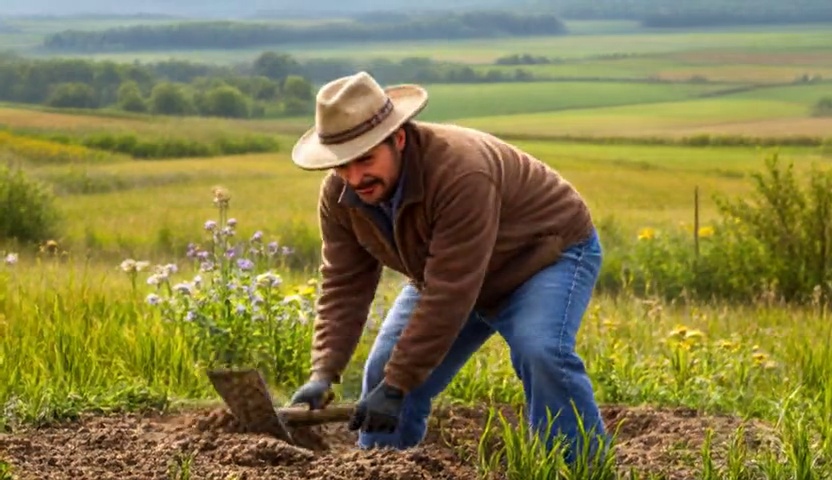}
    \hfill
        \includegraphics[width=0.23\linewidth]{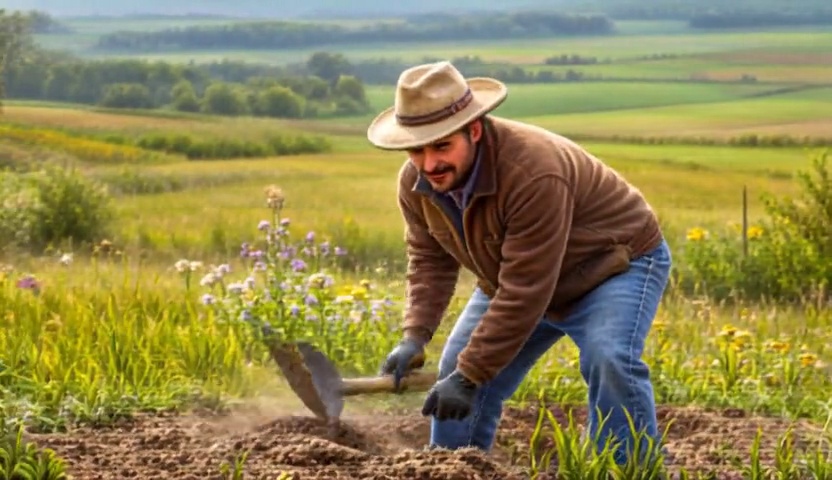}
    \hfill
    \includegraphics[width=0.23\linewidth]{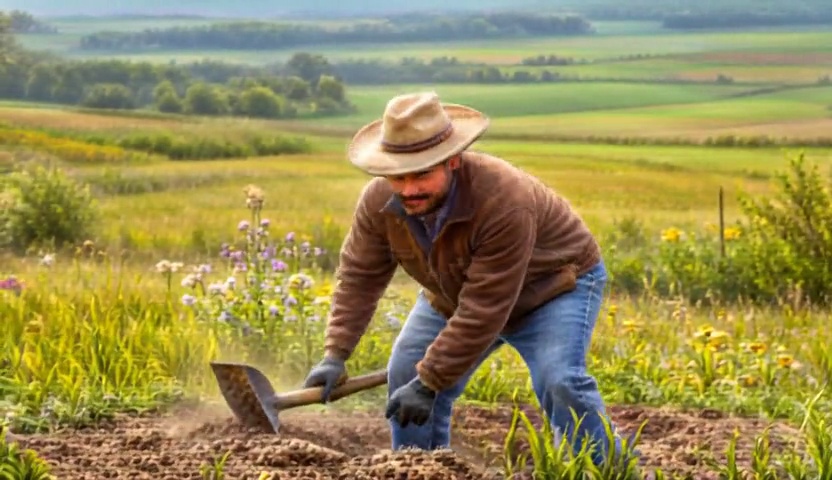}
    \subcaption{A person is digging}
    \end{subfigure}

    \begin{subfigure}[b]{\textwidth}
        \centering 
        \includegraphics[width=0.23\linewidth]{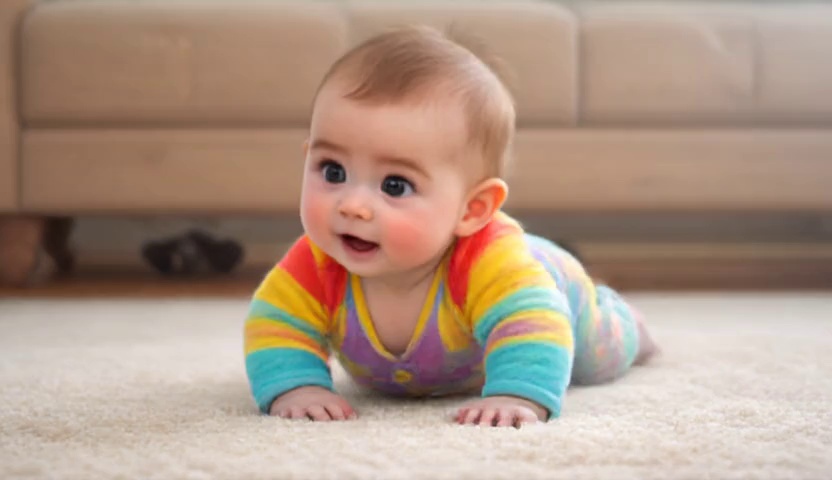}
        \hfill
        \includegraphics[width=0.23\linewidth]{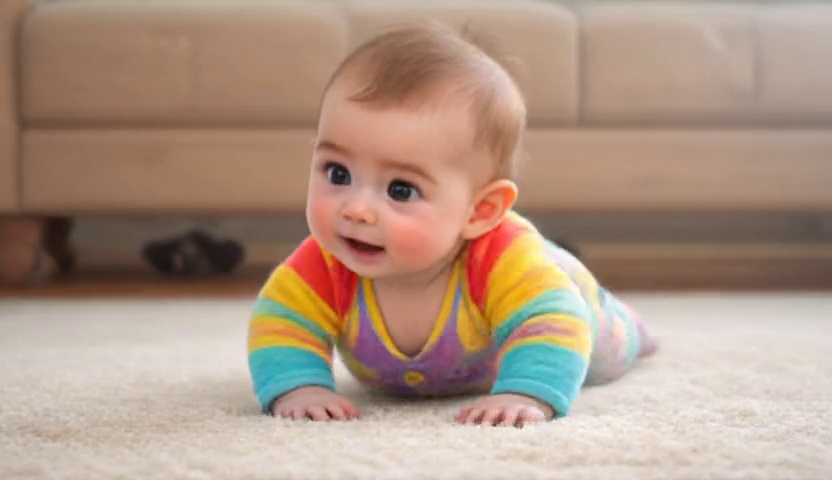}
        \hfill
        \includegraphics[width=0.23\linewidth]{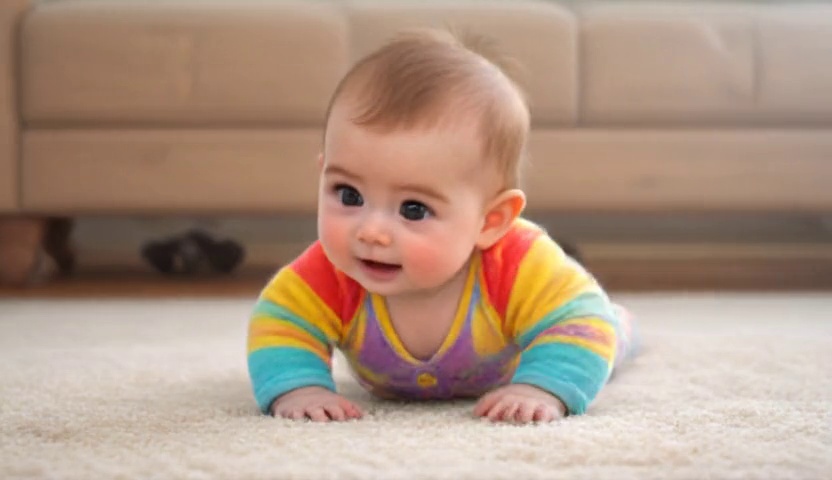}
        \hfill
        \includegraphics[width=0.23\linewidth]{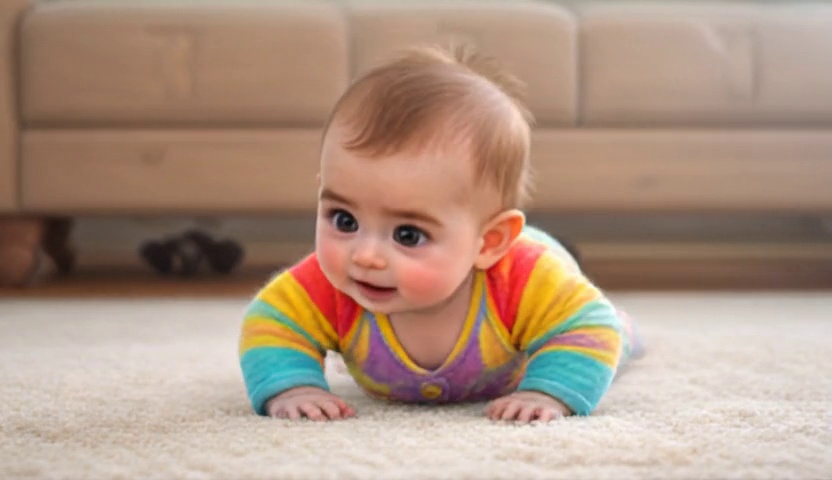}
        \subcaption{A person is crawling baby}
    \end{subfigure}

\begin{subfigure}[b]{\textwidth} 
    \centering 
    \includegraphics[width=0.23\textwidth]{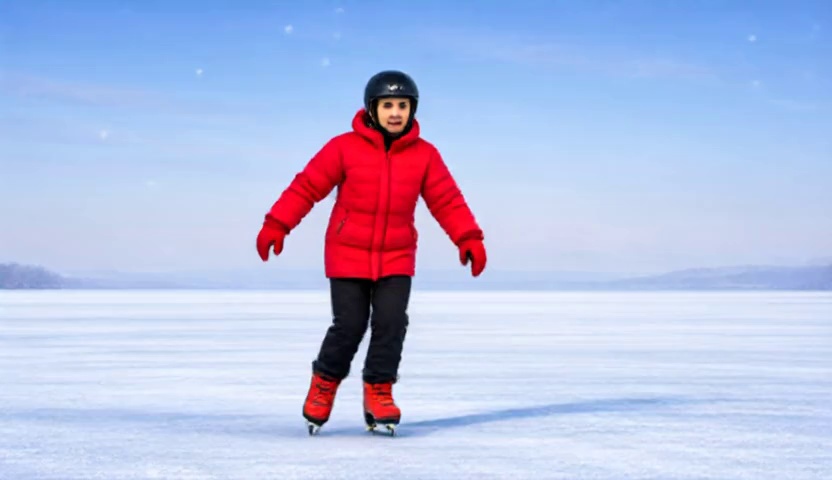}
    \hfill
    \includegraphics[width=0.23\textwidth]{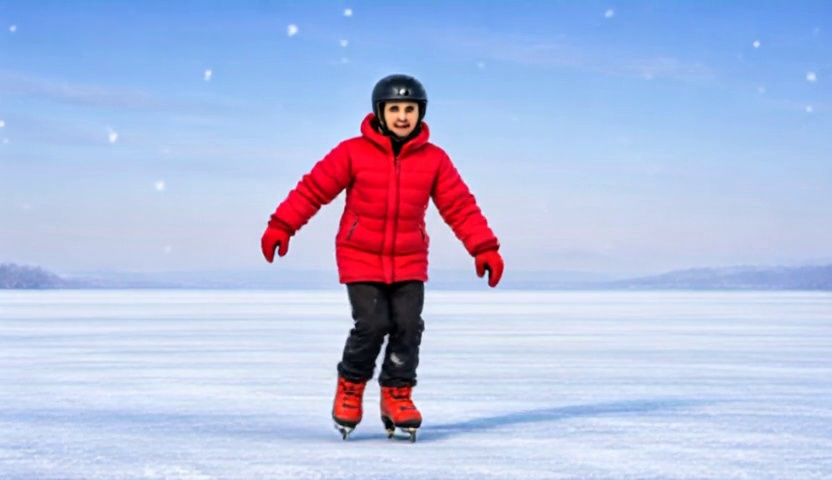}
    \hfill
    \includegraphics[width=0.23\textwidth]{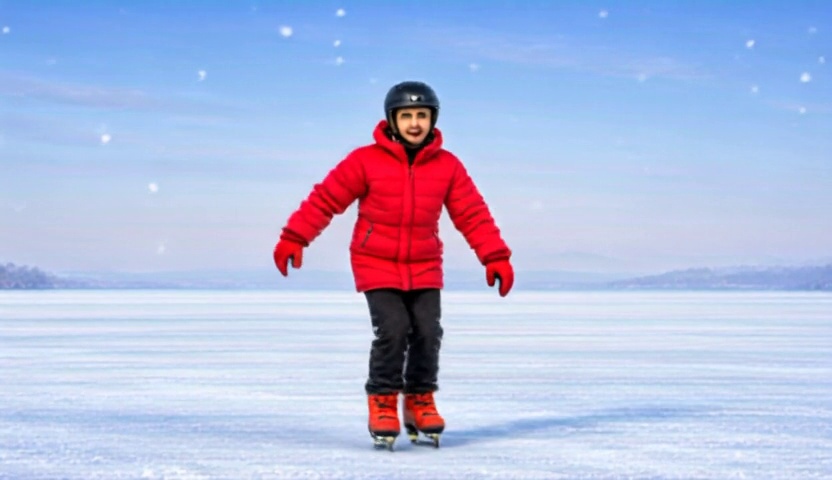}
    \hfill
    \includegraphics[width=0.23\textwidth]{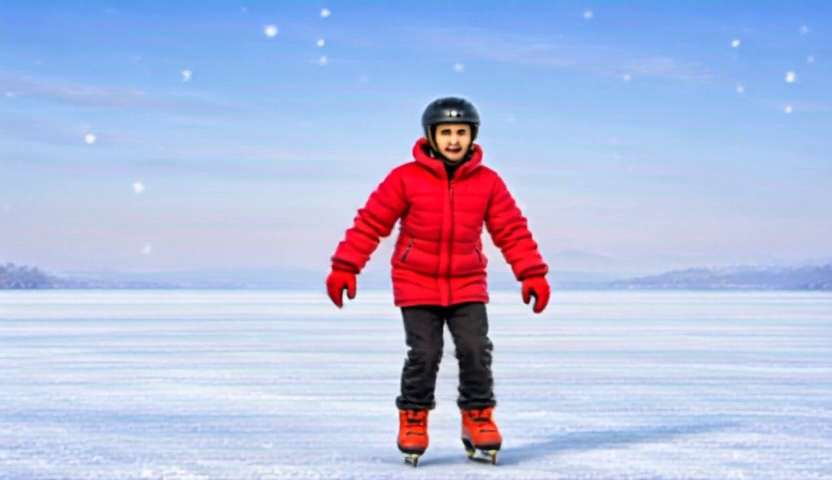}
    \subcaption{A person is ice skating} 
\end{subfigure}

    \caption{More generated samples. Prompts are randomly sampled from VBench of various scenes to benchmark the overall capability of the model.}
\label{fig:two-rows-eight-images-more}
\end{figure*}

\end{document}